\documentclass[journal]{IEEEtran}
\usepackage{cite}
\ifCLASSINFOpdf
\else
\fi 
\usepackage{multirow}
\usepackage{flushend}
\usepackage{graphicx}
\usepackage{colortbl}
\graphicspath{{Figures_eeg/}}
\usepackage{caption}
\usepackage{capt-of}
\usepackage{balance}
\usepackage{subcaption}
\usepackage{fixltx2e}
\usepackage{setspace}
\usepackage{amsmath}
\usepackage{hyperref}
\usepackage{array}

\begin{document}
\title{T-BERT - Model for Sentiment Analysis of Micro-blogs\\ Integrating Topic Model and BERT} \author{Sarojadevi  Palani, Prabhu Rajagopal, and Sidharth Pancholi,~\IEEEmembership{Member,~IEEE}

\thanks{ Sarojadevi Palani is with Center for Nondestructive Evaluation and Department of Mechanical Engineering, Indian Institute of Technology, Madras, India (e-mail: saroojadevi@gmail.com)}
\thanks{Prof.Prabhu Rajagopal is with Center for Nondestructive Evaluation and Department of Mechanical Engineering, Indian Institute of Technology, Madras, India (e-mail: prajagopal@iitm.ac.in)}
\thanks{Dr.Sidharth Pancholi is with the Department of Electrical Engineering, Indian Institute of Technology, Delhi, India (e-mail: s.pancholi@ieee.org)}}

\maketitle

\begin{abstract}
\noindent Sentiment analysis (SA) has become an extensive research area in recent years impacting diverse fields including e-commerce, consumer business, and politics, driven by increasing adoption and usage of social media platforms. It is challenging to extract topics and sentiments from unsupervised short texts emerging in such contexts, as they may contain figurative words, strident data, and co-existence of many possible meanings for a single word or phrase, all contributing to obtaining incorrect topics. Most prior research is based on a specific theme/rhetoric/focused-content on a clean dataset. In the work reported here,  the effectiveness of BERT(Bidirectional Encoder Representations from Transformers) in sentiment classification tasks
from a raw live dataset taken from a popular microblogging platform is demonstrated. A novel T-BERT framework is proposed to show the enhanced performance obtainable by combining latent topics with contextual BERT embeddings. Numerical experiments were conducted on an ensemble with about 42000 datasets using NimbleBox.ai platform with a hardware configuration consisting of Nvidia Tesla K80(CUDA), 4 core CPU, 15GB RAM running on an isolated Google Cloud Platform
instance. The empirical results show that the model improves in performance while adding topics to BERT and an accuracy
rate of 90.81\% on sentiment classification using BERT with the proposed approach.\end{abstract}





\section{Introduction}

\IEEEPARstart{T}{he field of Sentiment Analysis (SA)} studies opinions expressed in terms of micro-blogs, texts, people’s interest or opinion towards a topic, about product reviews, government policies, religion, etc. in natural human language  \cite{Medhat2014,Sharma2020,Hasan2018,Moshkin2020a}.

\begin{figure*}[htbp]
	\centering
	\includegraphics[scale=0.40]{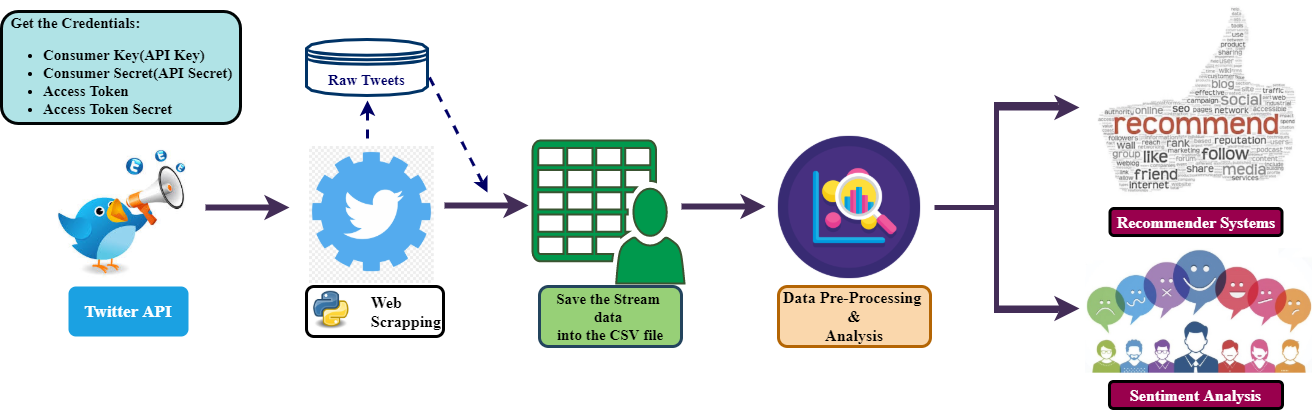}
	\centering \caption{Overall approach of proposed T-BERT}
	\label{OVERALL workflow} 
\end{figure*}

Businesses seek to use these contents in real time to understand how customers perceive their products and formulate strategies accordingly. This information can also be used to resolve product restrictions or monitor people’s attitudes before new products are released  \cite{Naseem2020}. The proliferation of these and also the advancements in Natural Language Processing (NLP) has provided researchers in giving more solutions to many engineered problems related to human language focusing on people emotions and opinions. The explosion of online social networking \cite{Hasan2018} on daily basis in turn demands the need of high Cloud computing Graphics Processing Unit (GPU) and Tensor Processing Unit (TPU) processors with a specialized purpose and architecture to store, process and analyse them\cite{Li2019,Balaji2017,Alaoui2019,Leung2016,Dwivedi2017}. However, the critical part lies in mining people’s opinions, intents towards discussions, emotions, any hidden patterns or latent information in the form of any natural human-human readable canonical language. Also, helpful data is covered in disarranged, inadequate, and unstructured instant messages. Some researchers, propose to arrange a lot of these messages or micro-blogs into groups with significant bunch marks, along these lines give an outline of the substance to address clients data needs \cite{Hu2015,hu2015embracing}. Statistical topic models fail to derive these patterns accurately when it comes to large and heavy-tailed corpus texts \cite{Cheng2014,Amit2007}. However, transformer based pre-trained models such as BERT (Bidirectional Encoder Representations from Transformers)   overcomes the need of supervised learning pattern. The transformers-based models \cite{Naseem2020} use a pre-trained sentence embedding vectors which is trained on large corpora of data so that the problem of supervised learning is not highly essential in here. It has been argued that, today's techniques limit the power of pre-trained representations, especially for fine-tuning approaches due to the restriction imposed on the standard uni-directional language models. For example, in OpenAI GPT, where  left-to-right architecture is used for pre-training ends up in looking for only the previous tokens in the attention layers of the transformers\cite{Devlin2019,Moshkin2020a}. This could be an acceptable approach while taking sentence level tasks but while considering token level tasks such as question answering, the fine-tuning tasks gets crucial as it involves processing from both the  directions\cite{Moshkin2020a}. In this research, a flexible framework to combine latent topic information with BERT embeddings \textbf{(T-BERT)} to extract the contextual topics from live twitter dataset is proposed \and to classify the sentiments using BERT on the same. The main contributions of this proposed approach are a) To build and develop modelS to extract latent topics b) To develop BERT sentence embeddings to draw the contexts based on the semantic similarity in the microblogs and merge them with LDA latent topics using a deep learning auto-encoder. c) To propose and build a BERT Sentiment classifier and to finally merge both contextual topics and sentiments for each microblog using software engineering techniques. Some of the traditional latent topics extraction methods plays excellent role in the areas of Information Retrieval (IR), Cyberbullying, analysing political texts from newspaper articles, large image content extraction, summarization etc. This study could favour some real time applications as shown in Fig~\ref{OVERALL workflow} where the Topic/Aspect-based Sentiment analysis is taken advantage of: 1. Recommendation systems 2. Market Research 3. Brand Monitoring 4. Earth quake detection algorithms 5. Financial analysis 6. Bio-informatics 7. Social Network Analysis (SNA) 8. Software engineering. The rest of the paper is structured as follows. Section II Related Work walks through several Sentiment analysis, Topic models, BERT based recent research papers. Section III; Explains algorithms used and work methodology. Section IV; Explains the setup for numerical experiments and various model hyper-parameters utilized. Empirical results of the BERT contextual topics and BERT Sentiments are discussed in Section V, after which the paper concludes with directions for further work.
\hfill 
\section{Related Work}
\noindent Sentiment analysis on unstructured data such as micro-blogs or short-texts or other inherently sparse data are conducted on various standard data sets where the data is collected under specific theme or topics or politics or movie reviews or using popular SemEval datasets \cite{Chen2018,Jianqiang2018,Abdi2019,Naseem2020,Ray2020}. However, in this study the theme or topics or genre is unknown in the microblogs. \cite{Alsaeedi2019} conducted a theoretical study and comparison on the 3 levels of SA tasks on short texts and gave a comprehensive state-of-art results based on the algorithms implemented by multiple researchers.  In recent years, research on analysing short-texts and micro-blogs has peaked  \cite{Cheng2014,Zou2018,Zheng2020} due to the mass usage of smart gadgets and social media platforms. \cite{Zou2018} conducted numerical experiments on Health Care Reform and United States Presidential election debates in the year 2009 by taking polarity labeled datasets and proposed a novel SA on micro-blogs incorporating homophily \cite{Ren2013} in social networks, additionally introducing topic context to cater the semantic relations between them. In 2007, \cite{Amit2007,Aziz2018} presented a sentiment polarity based TM to analyze public satisfaction at the government in Surabaya city. A methodology based on “Markov chain” was proposed by \cite{Amit2007} where all the words in the same sentence and its consecutive sentences falls under same topics and the model was trained and inferred using Hidden Markov tools. A comparative study on various TM techniques and Tool-kits from 2015 to 2020 evaluating performance metrics on major application areas with an excellent high-level report are presented in 
\cite{Albalawi2020,Li2016,Gao2019,Dieng2020,Hu2020}.

With the advancements in the Deep Learning Neural Networks and the increase in the demand for many Natural Language Processing(NLP) tasks from various domains and business intakes have led researches to make much deeper analyses and  helped develop the Transformers based architectures on top of DL networks which could solve many NLP tasks in a much efficient and faster way incorporating attention mechanism for robust  training \cite{Sutskever2014}. \cite{Vaswani2017} introduced a novel Transformer based architecture that uses the same encoder-decoder network similar to seq2seq model with attention mechanism but does not connote any LSTM, GRU or RNN networks. The Transformer architecture model completely relies on Attention mechanism thereby avoiding recurrence to draw global dependencies between the input and the output which in turn allows parallelism. In late 2018, inspired by the Transformers architecture Google \cite{Devlin2019} research team introduced a new deep Bidirectional Encoder Representations from Transformers called BERT to alleviate the constraints imposed on unidirectional methods.  Sentiment analysis
\cite{Hoang2019,Munikar2019,Zheng2020} outperformed with improved F1 scores while pre-training with BERT classifiers. \cite{Munikar2019} performed a sophisticated fine-grained sentiment analysis on publicly available popular Stanford Sentiment Treebank (SST) dataset. A new algorithm formulating specific training dataset for neural networks using Word2Vec and Bert to detect sentiment values was introduced by \cite{Moshkin2020a}. Numerical experiments conducted on Aspect-Based Sentiment Analysis (ABSA) \cite{Hoang2019,Song2020}
by fine-tuning the intermediate layers of BERT in the context of semantic detection between two sentences 
\cite{Song2020}. BERT \cite{Devlin2019} performance on Semantic Textual Similarity(STS) tasks such as sentence-pair regression had proven to be a state-art for many NLP tasks but requires both the sentences to be fed to a cross-encoder transformer network simultaneously to predict the output \cite{Reimers2020}. In this study, SBERT(Sentence-BERT) introduced by \cite{Reimers2020} uses a different architecture called Siamese and triplet network thereby comparing the semantically meaningful sentences using cosine-similarity drastically reducing 65 hours of  work load to $\approx$5 seconds, with no compromise in accuracy from BERT is utilized for generating sentence embedding vectors to ease computational power.  
\section{METHODS}
\noindent Inspired by BERT and the recent advancements in Transformers architecture \cite{Moshkin2020a} \textbf{T-BERT} framework is proposed to show improved performance on topic model and aims at predicting the sentiment polarity (positive, negative or neutral) of the microblogs in study. \subsection{DATASET AND PRE-PROCESSING}
\noindent Data is extracted from popular micro-blogging platform Twitter via the link \url{https://apps.twitter.com/} using Tweepy API \cite{lwowski2018geospatial} credentials, python web scrapping techniques and libraries without any search keywords or specific topics of interest.  In this study, around 40k raw microblog without any user information are extracted into a spreadsheet and processed for data analysis. The following pre-processing steps are performed before being fed into model algorithms. 1. Canonicalization (Removal of numbers, Non-Ascii characters, punctuation symbols, accents, and convert microblogs to lowercase.) 2. Removal of URL links 3. Removal of Stop words 4. Replace emoticons with their original text form 5. Stemming/Lemmatization – Converting to root words 6. Tokenization of microblogs using tokenizer specific to BERT input and spacy libraries for LDA topics.
\begin{eqnarray} \label{eq2}
\nonumber \noindent P(W,Z,\theta,\phi  \mid \alpha,\beta)  = \end{eqnarray} \vspace{-1cm} \begin{flushleft} \begin{eqnarray}
   \prod_{j=1}^{M} P(\theta_j; \alpha)  \prod_{j=1}^{K} P(\phi_i; \beta) {\prod_{t=1}^{N} P(Z_{j,t} \mid \theta_j), P(W_{j,t} \mid \phi Z_{j,t}) }
\end{eqnarray} \end{flushleft} \vspace{-1cm} \begin{figure}[h]
	\centering
	\includegraphics[width=\linewidth]{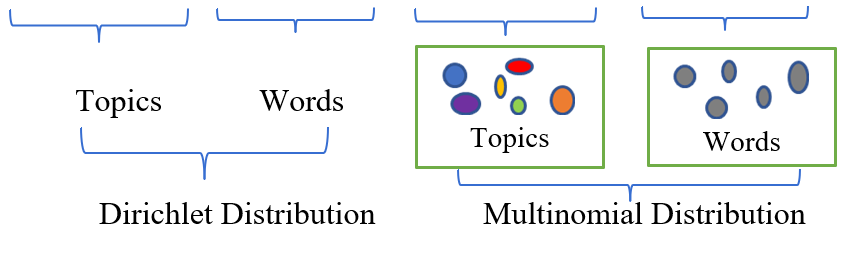}
	\end{figure}

\begin{figure*}[htbp]
	\centering
	\includegraphics[width=\linewidth]{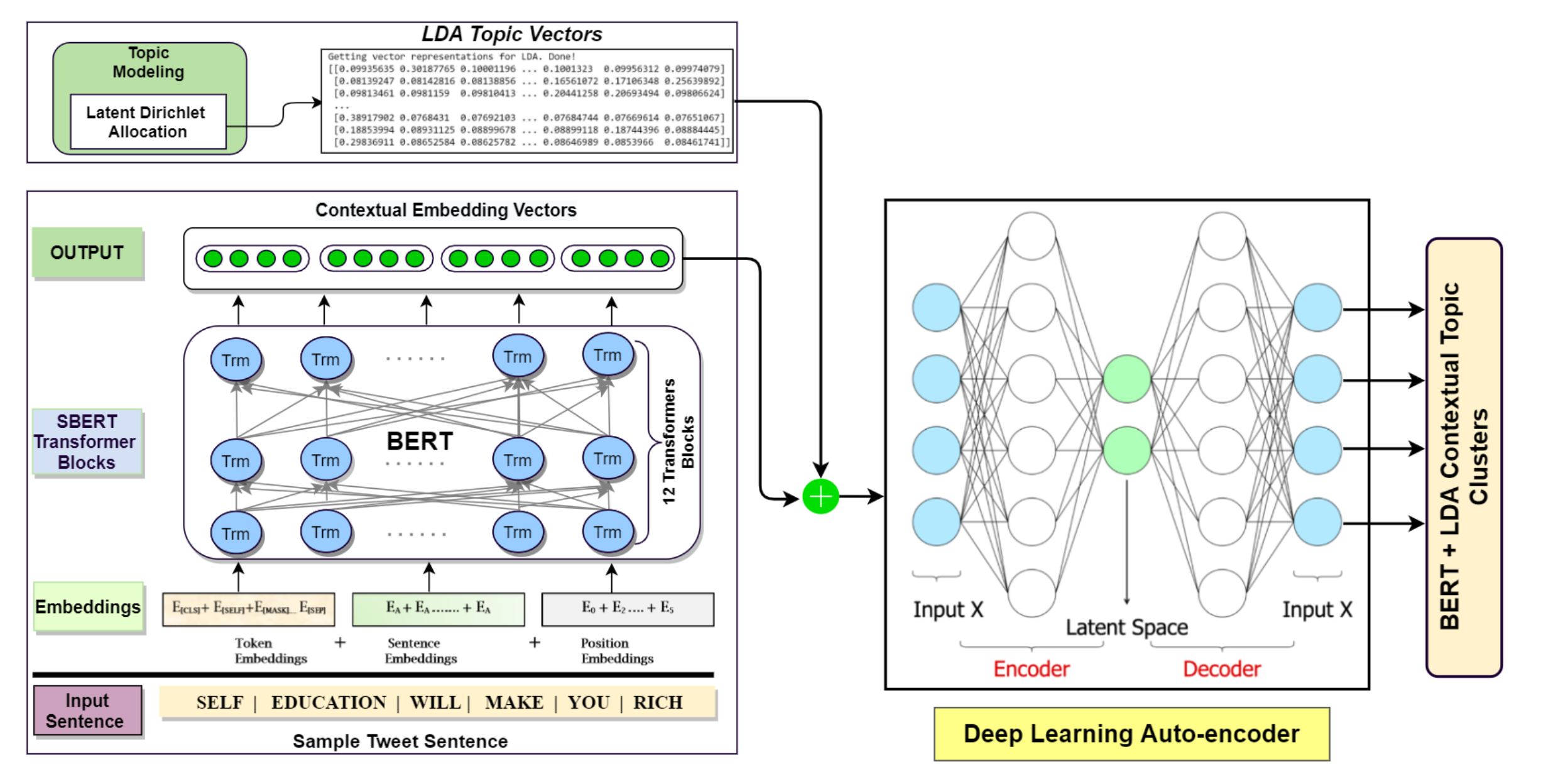}
	\centering \caption{Detailed workflow of BERT + LDA using deep learning auto-encoder.}
	\label{detailed process workflow} 
\end{figure*} \vspace{-0.5cm}
\subsection{LDA + BERT CONTEXTUAL TOPICS}
\noindent LDA adopts a general probabilistic approach to model rich corpora of data. Each tweet as input is represented as random mixture of latent topics (z), where each topic is a collection or distribution of words (w) related to topics. The goal of LDA is to arrive at a probabilistic model for a corpus of documents that assigns high probability to the members of the documents and also assigning high probability to other “similar” documents in the corpus. The model algorithm works as to arrive at the topic’s distribution as referred in the expanded equation in \eqref{eq2}: Thus, $\alpha$ is the parameter of the Dirichlet prior on the per-document topic distributions, $\beta$ is the parameter of the Dirichlet prior on the per-topic word distribution. The latent topic words are limited by the traditional structure of the LDA bag-of-words and PoS tagging model, which cannot effectively compare the semantic similarities in the tweet sentences and retrieve contextual information of the text. Perhaps, BERT could alleviate this when combined with LDA and a detailed process workflow of the same is shown in Fig.~\ref{detailed process workflow}. 
In order to obtain the contextual topics, the topic vectors ($\omega$) of LDA model are merged with a collective contextual word embedding vectors (H) from Sentence-BERT model using a gamma ($\gamma$) hyper-parameter to add relative importance to both vectors as shown in \eqref{eqn3}. $H = BERT ({x_1}, . . ., {x_T})$	
\begin{equation} \label{eqn3}
    \textit{Contextual Topic Vectors} (t) = \omega\gamma + H 
\end{equation}
where ${x_1, . . ., x_T}$ is a collective word vector of word embedding, segment embedding, and position embedding of each tweet token; Trm stands for Transformer encoder unit; $H = (w_1, . . ., w_T)$, and $w_i$ the averaged output from 12 multi-headed transformer blocks given as token’s contextual embedding vector representation. The combined vectors(t) are passed into a deep learning auto-encoder latent vector space to ensure dimensionality reduction and noise to arrive at the best topic clusters. The output of the auto-encoder is a cluster of keywords, each falling into a specific unique topic category using a K-Means clustering algorithm are labeled manually with unique topics. This approach is expected to provide more accurate topic semantic information for simulation on short tweet texts.\subsection{BERT Sentiment Classification}
\begin{figure*}[!h]
	\centering
	\includegraphics[width=\linewidth]{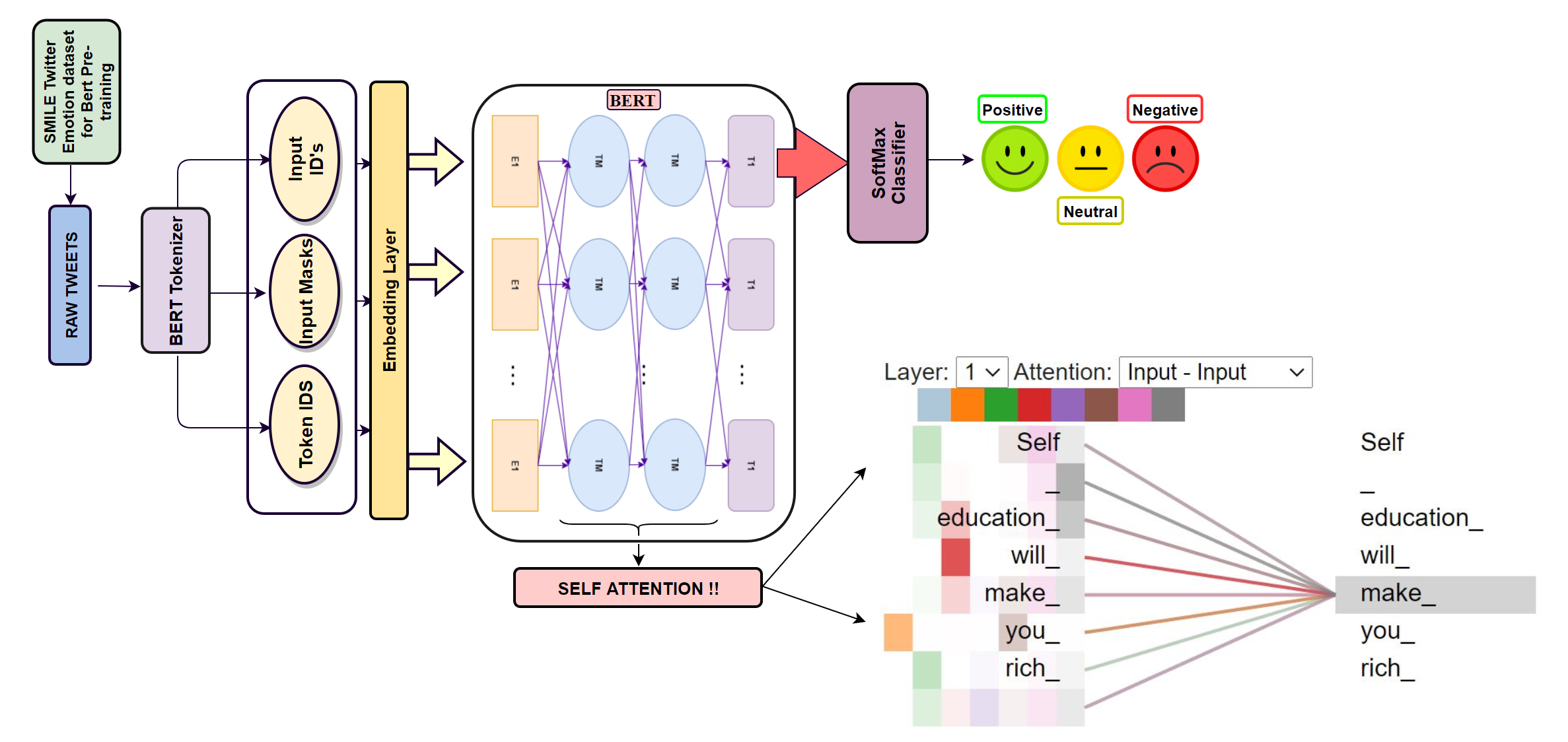}
	\centering \caption{Workflow of BERT Sentiment Analysis detailing Self Attention}
	\label{SA workflow} 
	\end{figure*}

\noindent The sentiment analysis model in Fig~\ref{SA workflow} aims to detect accurately the emotions of the microblogs, grasp the semantic characteristics and rules of the emotional evolution, and assist the contextual topics by adding value to the business in terms of recommendations, aspect-based sentiments, understand user intents to post relevant and appropriate ads etc. For the purpose of BERT pre-training, SMILE Twitter Emotion dataset adapted from \url{https://figshare.com/articles/dataset/smile_annotations_final_csv/3187909} is utilized and later the model is predicted for sentiments classes using twitter dataset used for topics extraction in this study. Firstly, SMILE Twitter emotion corpus is pre-trained in the BERT, after data pre-processing like Canonicalization, tokenization using Bert tokenizer and finally the [CLS] and [SEP] tokens are added at the appropriate positions by the model in itself. Secondly, the tokenized corpus is used to perform in-depth pre-training of the BERT target field on the constructed sentiment classifier. Fig~\ref{SA workflow} explains the attention mechanism showing the weights of each word but the other words in the tweet sentence and the words with contextual importance is highlighted in darker color for each layer of the transformer blocks.  For example, the word ‘Self’ carries importance when the word ‘you’ is read from left to right from which the model learns the semantic similarities of the sentence thus providing more accurate meaning behind the context, grammar and sentiments being discussed. In order to learn semantic information from the twitter texts, the transformer encoder connects the multi-headed self attention and feed-forward through a residual network structure. The multi-headed mechanism performs multiple linear transformations on the input vector to obtain different linear values, and then calculates the attention weight as shown in \eqref{eqn4} and \eqref{eqn5}.

	\begin{equation} \label{eqn4}
	 \textit{MultiHead (Q, K, V)} = Concat (h_1, h_2, . . ., h_T) W^M	
 	\end{equation}
	\begin{equation} \label{eqn5}
	HF = Attention (QW^Q, KW^K, VW^V)
	\end{equation}
	
where Q, K, V are the input word vector matrix of queries, keys and values respectively. Attention is calculated after Q, K, and V are mapped through the parameter matrix, and the calculation results are spliced after repeating t times. $W^M$ is the weight matrix. $W^Q$, $W^K$, $W^V$ represents the weight matrix corresponding to the $T^{th}$ hyper-parameter head. Thus, the Transformer encoder learns and stores the semantic relationship and grammatical structure information of the microblogs and based on this input the SoftMax classifies the sentiments easily. 

\section{SETUP FOR NUMERICAL EXPERIMENTS}
\noindent All numerical experiments were conducted with BERT$_{BASE}$(uncased) with 12 transformer blocks, 768 hidden units, 12 self-attention heads and overall 110M trainable parameters. Out of 42,626 raw tweets 23,329 tweets after pre-processing are fed to LDA and Bert for contextual topic assessment and as test dataset for sentiments predictions.

\subsection{EVALUATION METRICS}
\noindent The following metrics are used for Sentiment analysis
\begin{equation} \label{eqn6}
\textit{\textbf{1)	F1-Score}} = 2 \times \frac{(precision \times recall)}{(precision + recall)} \in (0,1)
\end{equation}   

\begin{equation} \label{eqn7}
\textit{\textbf{2)	Accuracy}} =   \frac{(TN+TP)}{(TN+TP+FN+FP)} \in (0,1) 	
\end{equation}
Where TP is True Positive, FP is False Positive and FN is False Negative. For topic model coherence scores measures and for Bert embeddings K-Means clustering Silhouette metrics are considered.

\begin{equation} \label{eqn8}
 CoherenceScore= \sum_{i<j} score(w_i,w_j) 
 \end{equation}

\subsection{\textbf{Model Building and Hyper-parameters}}
\begin{figure}[htbp]
	\centering

	\includegraphics[width=\linewidth]{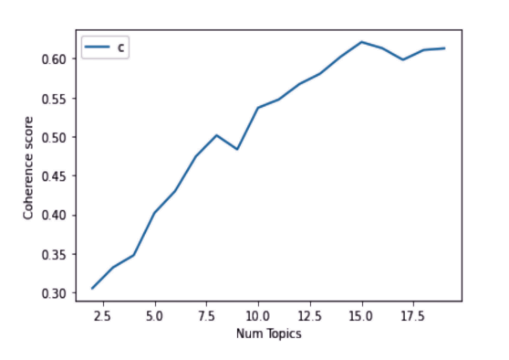}
	\centering \caption{Optimal Coherent model output showing (k) value 8 as cut-off point}
	\label{CUT-off Scores} 
\end{figure}

\noindent In Engineering, LDA is set to automatically classify documents and estimate their relevance to various latent topics hidden in them. In this study, document term matrix after pre-processing is achieved using genism doc2bow $($bag of words$)$ toolkit for a dictionary of 11057 unique words. Multiple numerical experiments were conducted on LDA with random values of ($\alpha$) alpha ranging between 0 $<$ $\alpha$ $<$ 1 (0.01, 0.1, 0.2 and 0.3) and the optimal topic model(k) is arrived as shown from Fig~\ref{CUT-off Scores}.
\\ \hfill
\noindent \textbf{Autoencoder Hyperparameters:}\\
The list of hyper parameters to fine-tune auto encoder are:

    \begin{itemize}
      \item Batch size
      \item Epochs
      \item Kernal\_regularizers (L1 and L2) to add weights to the layers.
      \item Dropouts to control over fitting.
      \item Learning Rate.
      \item Gamma parameter to control the relative   importance of both LDA and Bert vectors.
    \end{itemize}
Extensive numerical experiments were conducted with various hyper-parameters for fine tuning the auto-encoder layers for topics ranging from 5 – 17 are displayed in Table \ref{tab:1} to compare the performance on each (k) value.

\begin{table*}[!h]
\centering
\caption{Summary of Auto-encoder Hyper-parameters used in LDA+BERT numerical experiments}
\label{tab:1}
\begin{tabular}{|c|c|c|c|}
\hline
\multicolumn{4}{|c|}{\textbf{Topics K   (5,6,7,8,10,12,15,17), Optimizer = Adam, Gamma = 15}}           \\ \hline
\textbf{Epochs} & \textbf{Batch size} & \textbf{Regularizers}                                                                                  & \textbf{Dropout Rate} \\ \hline
50  & 128 & L1(10e-4) & 0.01  \\ \hline
100 & 128 & L1(10e-4) & 0.01  \\ \hline
200 & 128 & L1(10e-4) & 0.01  \\ \hline
300 & 128 & L1(10e-2) & 0.001 \\ \hline
50  & 128 & L2(10e-4) & 0.1   \\ \hline
500             & 128                 & \begin{tabular}[c]{@{}c@{}}Bias\_regularizer L2(1e-5), \\ Activity\_regularizer L1(10e-2)\end{tabular} & 0.001                 \\ \hline
\end{tabular}
\end{table*}

\begin{table}[htbp]
\centering
\caption{Experimental results of LDA and BERT Coherence scores for each topic(k) value in (\%)   
}
\label{tab:Table 2}
\begin{tabular}{|c|l|l|}
\hline
\textbf{\# of Topics(k)} & \multicolumn{1}{c|}{\textbf{LDA}} & \multicolumn{1}{c|}{\textbf{BERT}} \\ \hline
5  & 0.402 & 0.481 \\ \hline
6  & 0.430 & 0.490 \\ \hline
7  & 0.474 & 0.451 \\ \hline
8  & 0.501 & 0.521 \\ \hline
10 & 0.483 & 0.504 \\ \hline
12 & 0.537 & 0.459 \\ \hline
15 & 0.567 & 0.441 \\ \hline
17 & 0.621 & 0.460 \\ \hline
\end{tabular}

\end{table} 
\begin{figure*}[hbpt]
	\centering
	\includegraphics[width=\linewidth]{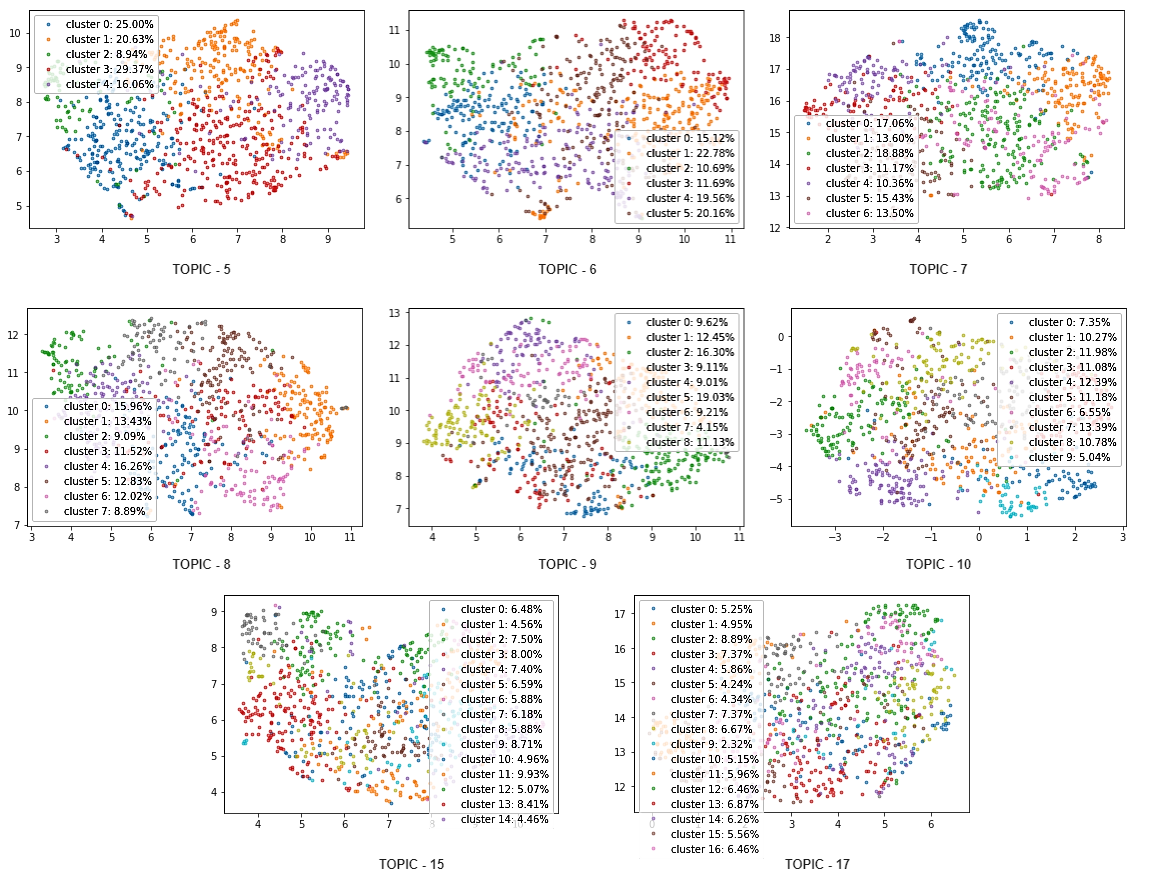}
	\centering \caption{Visualization of BERT Clustering experimented for varying number of topic values (5,6,7,8,10,12,15,17)}
	\label{Clustering} 
\end{figure*}

\section{RESULTS and DISCUSSION}
\noindent In this section, the empirical results of the numerical experiments and related findings are evaluated to assess the effectiveness of the proposed framework \textbf{T-BERT}. In particular the results of LDA are evaluated against BERT and LDA+BERT. The metrics for each topic(k) shown in Fig~\ref{CUT-off Scores} clearly indicating a gradual rise in the coherence scores until the cut-off point, with k=8 being optimal value. The dataset being unsupervised, BERT contextual embedding results are clustered and visualized using K-Means clustering algorithm grouping similar topics as shown in Fig~\ref{Clustering}  and the evaluation metrics from Table \ref{tab:Table 2}.

\noindent The weighted method obtains better accuracy's in terms of coherence values adding contexts to the topics(k=8) which makes the clustering of similar topics look more meaningful than LDA topics. Adding weights to auto-encoder layers using L1 Regularizers resulted in over-fitting. However, a dropout of 0.01 and Adam optimizer \cite{Vaswani2017,Reimers2020,Song2020} at the input encoder layer randomly turned off some redundant layers in the network avoiding over-fitting and to ensure a robust model output displayed in Table \ref{tab:Table 3}. 

\begin{table}[htbp]
\centering
\caption{Final Auto-encoder Hyper-parameters}
\label{tab:Table 3}
\begin{tabular}{|l|c|}
\hline
\multicolumn{1}{|c|}{\textbf{Parameters}} & \textbf{Values} \\ \hline
Latent dimension                          & 64              \\ \hline
Activation Function                       & RELU            \\ \hline
Epochs                                    & 50              \\ \hline
Batch size                                & 128             \\ \hline
Regularizers                              & L1(10e-4)       \\ \hline
Gamma                                     & 15              \\ \hline
Dropout Rate                              & 0.01            \\ \hline
Optimizer                                 & Adam            \\ \hline
\end{tabular}
\end{table}

\begin{table*}[htbp]
\centering
\caption{Final Metrics (\%) for Contextual Topic Model (k = 8)
comparing LDA, BERT and BERT+LDA 
}
\label{tab:Table 4}
\begin{tabular}{|l|c|c|c|}
\hline
\textbf{Metrics / Scores} &
  \textbf{LDA} &
  \textbf{BERT + Clustering} &
  \textbf{\begin{tabular}[c]{@{}c@{}}LDA + BERT + \\ Clustering\end{tabular}} \\ \hline
\begin{tabular}[c]{@{}l@{}}C\_V \\ (Coherence scores)\end{tabular} &
  0.501 &
  0.521 &
  0.56 \\ \hline
Silhouette &
  NA &
  0.044 &
  0.46 \\ \hline
\end{tabular}
\end{table*}

\begin{figure*}[htbp]
     \centering
     \begin{subfigure}[b]{0.3\textwidth}
         \centering
         \includegraphics[width=\textwidth]{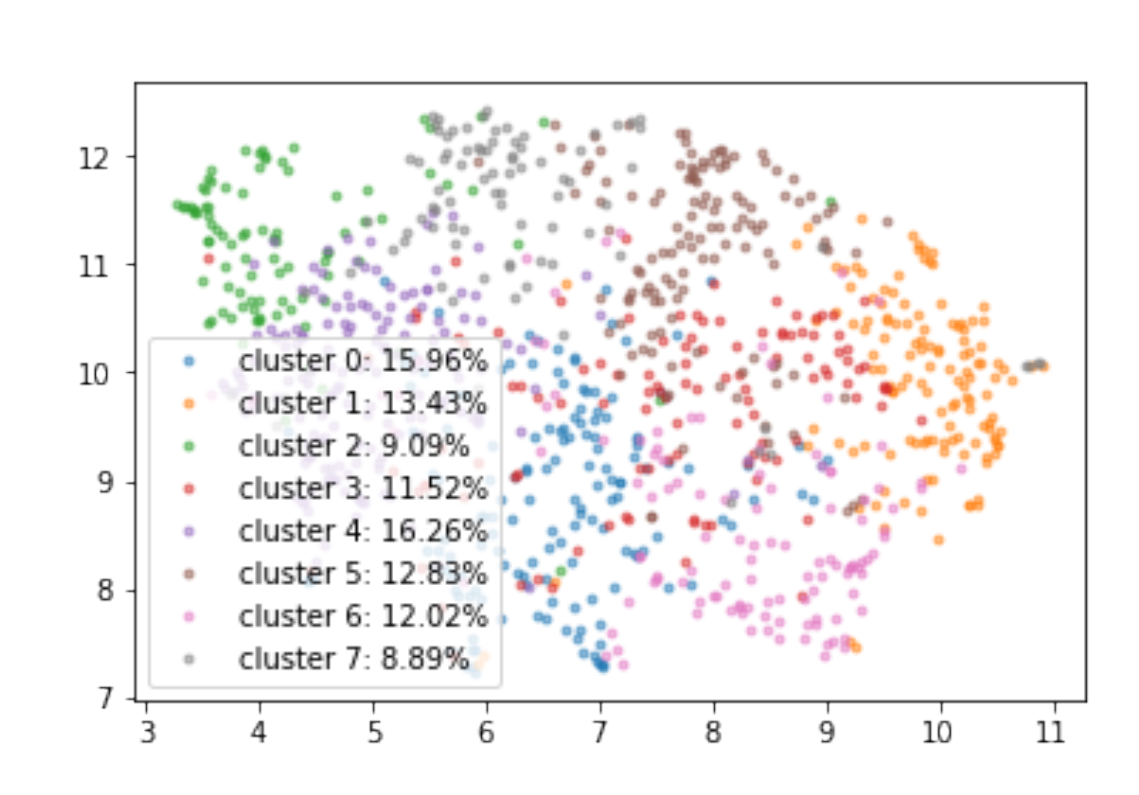}
         \caption{BERT}
         \label{fig:Fig16}
     \end{subfigure}
      \hspace{0.5em}
     \hfill
     \begin{subfigure}[b]{0.3\linewidth}
         \centering
         \includegraphics[width=\textwidth]{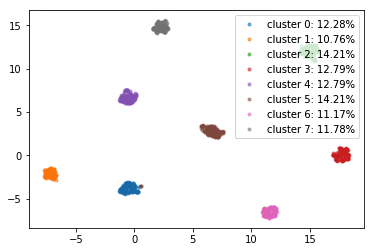}
         \caption{$LDA+BERT$}
         \label{fig:Fig17}
     \end{subfigure}
      \hspace{0.5em}
     \hfill
     \begin{subfigure}[b]{0.3\linewidth}
         \centering
         \includegraphics[width=\textwidth]{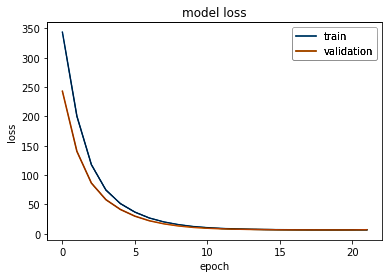}
         \caption{Training and Validation Loss of Auto-encoder}
         \label{fig:Fig18}
     \end{subfigure}
     \caption{Final Clusters for topic value k=8 along with Training and validation loss of Auto-encoder}
        \label{fig:Fig19}
\end{figure*}

\begin{figure*}[htbp]
	\centering
	\includegraphics[width=\linewidth]{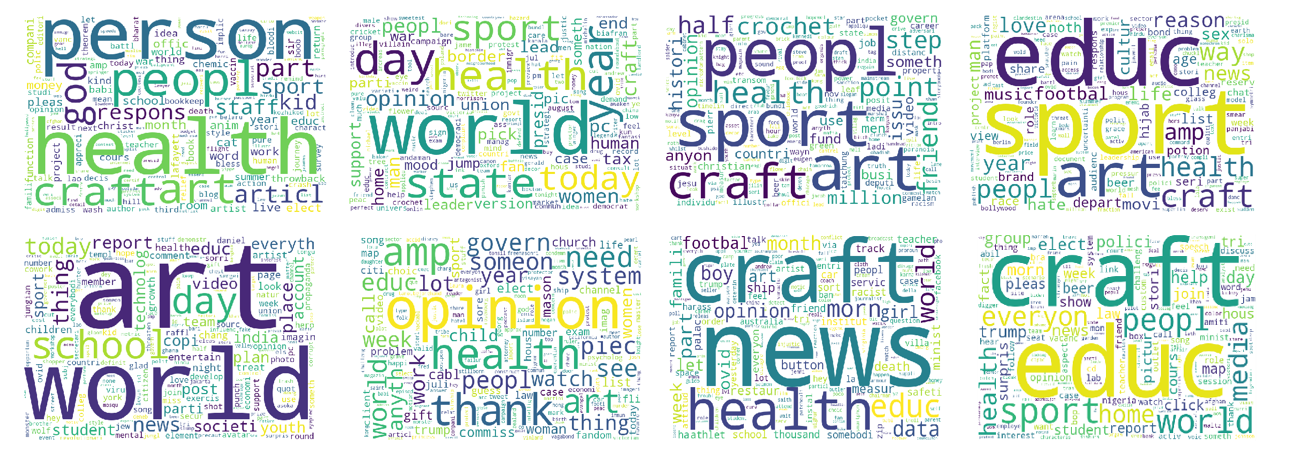}
	\caption{Visualization of word clouds after LDA and BERT merge. The contextual topic key words for optimal value k= 8 with each word cloud representing each topic clusters.}
	\label{WORD CLOUD} 
\end{figure*}

\begin{table}[!h]
\centering
\caption{BERT Hyper-parameters for Sentiments Classification}
\label{tab:Table 5}
\begin{tabular}{|l|c|}
\hline
Optimizer      & AdamW (Pytorch) \\ \hline
Epochs         & 10              \\ \hline
Batch size     & 32              \\ \hline
Learning rate  & 1e-5            \\ \hline
Adam's epsilon & 1e-8            \\ \hline
\end{tabular}
\end{table}
\begin{figure}[!h]
	\centering
	\includegraphics[width=0.75\linewidth]{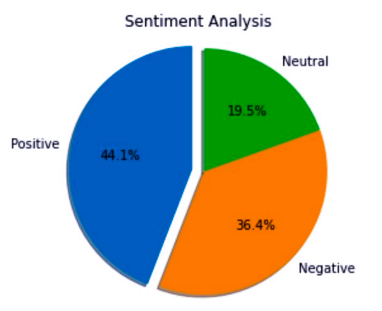}\vspace{-0.5mm}
	\centering \caption{BERT Sentiment predictions on test data-set}
	\label{Bert SA test results} 
\end{figure}
\begin{table}[!h]
\centering
\caption{Accuracy per class – Validation results}
\label{tab:Table 6}
\begin{tabular}{|l|c|}
\hline
\textbf{Class} & \textbf{Validation Accuracy (\%)} \\ \hline
Positive       & 0.96                              \\ \hline
Negative       & 0.78                              \\ \hline
Neutral        & 0.69                              \\ \hline
\end{tabular}
\end{table}

\noindent The goal of this research to demonstrate performance improvement by adding contextual sentence embedding of BERT with LDA topic model with proven accuracy improvements as shown from Table \ref{tab:Table 4}. Fig~\ref{fig:Fig16} shows the clustering results of BERT and Fig~\ref{fig:Fig17} the improved results of LDA+BERT showing a balanced and quite separated clusters. The gains between each model are relatively small, but the improvements of the method are reasonably straight forward and flexible which makes it easier to apply to a variety of other tasks. Visualization of word clouds for final 8 topics in Fig~\ref{WORD CLOUD} shows that each cluster highlights the proportionality and the frequency of words in similar topics are grouped together. The words bigger in size directly relates to its probability of the word occurrences within the topic. The goal for the numerical experiments on BERT sentiment analysis is two-fold. First, is to evaluate whether the training on the labeled is useful for validation on sentiments classifier on Smile twitter dataset. Second, to evaluate the performance and accuracy on testing the proposed twitter dataset and how well the sentiments are classified on the test data. The model architecture is built in such a way the data loader for both training and validation is sampled from a random and sequential sampler respectively with a batch size of 32 using pytorch pre-trained bert uncased model on a 12GB GPU machines with a fully connected layer. The summary of parameters used in the numerical experiments are shown in the Table \ref{tab:Table 5}. BERT significantly performs well in classifying the emotions in terms of accuracies (\%) as shown from Table \ref{tab:Table 6}. In this this study, BERT pre-trained model with its bi-directional and attention mechanism outperforms the predictions of emotions with its improved semantic search approach compared to other state-of-art models.

\section{CONCLUSION} 
\noindent The potential of leveraging the BERT attention mechanism is investigated in this work, and the recommended contextual subjects and sentiments underlying those topics (positive, negative, or neutral) are realized through the enhanced performance of T-BERT. Results from numerical experiments demonstrate the effectiveness and generality of the proposed approach. Future study could concentrate on merging direct topic information into BERT pre-trained models with more advanced ways to manage complicated data features, linguistic skills, and semantic search, as well as identifying attitudes in languages other than English.

\section*{Acknowledgment}
The first author would like to thank Liverpool John Moores University(LJMU), UK for approving this research work as part of her Masters thesis dissertation. 

\bibliographystyle{IEEEtran}
\bibliography{TBERT}

\end{document}